# Human-machine knowledge hybrid augmentation method for surface defect detection based few-data learning

Yu Gong[a], Xiaoqiao Wang[a], Chichun Zhou[b, c]

[a] *School of Mechanical Engineering, Hefei University of Technology, Hefei, 230009, China*
[b] *School of Engineering, Dali University, Dali, 671003, China*
[c] *Air-Space-Ground Integrated Intelligence and Big Data Application Engineering Research Center of Yunnan Provincial Department of Education*

**Abstract**

Visual-based defect detection is a crucial but challenging task in industrial quality control. Most mainstream methods rely on large amounts of existing or related domain data as auxiliary information. However, in actual industrial production, there are often multi-batch, low-volume manufacturing scenarios with rapidly changing task demands, making it difficult to obtain sufficient and diverse defect data. This paper proposes a parallel solution that uses a human-machine knowledge hybrid augmentation method to help the model extract unknown important features. Specifically, by incorporating experts' knowledge of abnormality to create data with rich features, positions, sizes, and backgrounds, we can quickly accumulate an amount of data from scratch and provide it to the model as prior knowledge for few-data learning. The proposed method was evaluated on the magnetic tile dataset and achieved F1-scores of 60.73%, 70.82%, 77.09%, and 82.81% when using 2, 5, 10, and 15 training images, respectively. Compared to the traditional augmentation method's F1-score of 64.59%, the proposed method achieved an 18.22% increase in the best result, demonstrating its feasibility and effectiveness in few-data industrial defect detection.

## 1. Introduction

Quality control is a critical technology in the manufacturing process that ensures products meet required standards [1,2]. In recent years, deep learning techniques have made significant strides in visual quality inspection tasks [3,4] due to the massive growth of manufacturing data and the powerful computing power of GPUs. This method possesses strong feature extraction and representation abilities for underlying data and complex structures. Additionally, its rapid adaptability to new products makes it well-suited to solving quality control issues for products on flexible production lines [5]. However, the data-driven method requires a sufficient amount of high-quality supervisory data. In actual industrial production scenarios, the defect rate is low, and abnormal data is scarce or non-existent, making data collection time-consuming and expensive [6]. As a result, supervised learning models perform poorly due to the lack of data [7], and the few-data problem has become a bottleneck preventing the practical implementation of defect detection in many cases.

In this situation, the classical solution is to use data augmentation methods [8–10] to expand existing data through geometric deformation such as translation, rotation, scaling, and mirroring. One or more deformations are applied to the data while preserving the semantic label during the transformation process [11]. However, this method is not designed for enhancing specific features and does not add any new practical additional information. The limited repetitive operations will saturate after a certain point [12], and the model becomes more sensitive to the changes brought by simple transformations, even leading to overfitting.

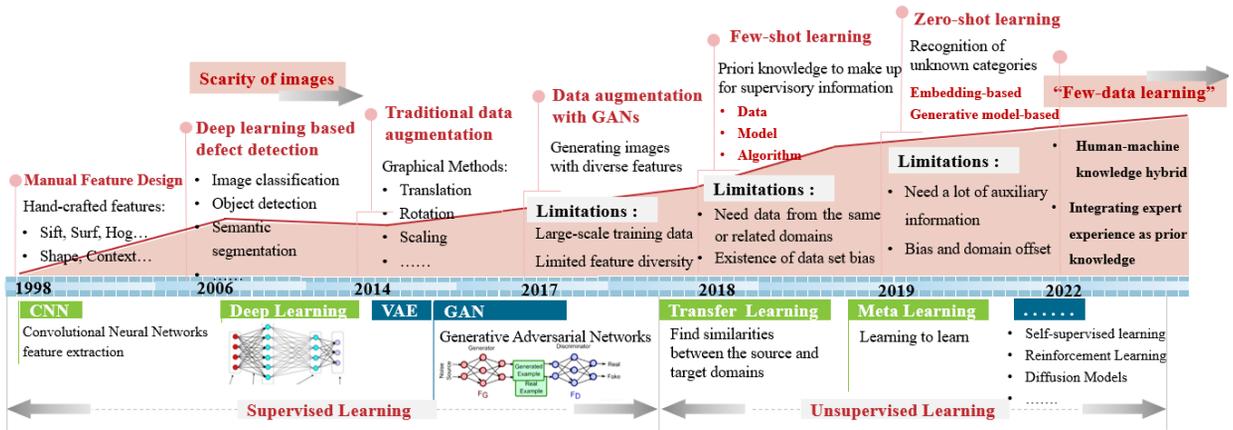

Fig. 1. Development schematic of industrial defect detection methods.

In recent years, an increasing number of researchers have utilized learning-based methods to overcome the issue of inadequate training data [13]. The most popular method involves the use of generative adversarial networks (GANs) for data generation. Several experiments have demonstrated that GANs can generate a large number of defect images with high fidelity and diversity [14,15], facilitating the effective training of deep defect detection networks. Consequently, numerous studies have explored the use of advanced generation models for data augmentation. However, utilizing GANs to generate defect samples presents several challenges. Existing GANs still require a substantial amount of training data, which does not adequately address the problem of limited data. Additionally, although GANs can enhance the effectiveness of existing data, they may generate only simpler structures and patterns [16]. The generated defect features exhibit singularity in attributes such as position, direction, and shape, making them unsuitable for industrial environments with varying conditions. These limitations considerably constrain the performance of subsequent automatic detection algorithms.

In the literature, few-shot learning (FSL) is also a typical method for addressing the issue of insufficient labeled data. FSL mainly uses prior knowledge to make up for the lack of supervision information. Current solutions include transfer [17], unsupervised learning [18], meta-learning [19], etc. FSL method can generally be classified into three categories: enhancing training data as prior knowledge [20], improving the model to limit the hypothesis space more effectively [21], and improving the algorithm to find the optimal hypothesis in the given hypothesis space [22]. These methods have been widely used in many fields, such as object detection [23], image segmentation [24], and image classification [25], and there have been successful examples in defect detection [26,27].

Following FSL, there have been zero-shot learning (ZSL) solutions. The basic idea of ZSL is to use some visible class data, supplemented with relevant information or prior knowledge as attribute labels, to identify unknown classes [28,29]. Current ZSL can generally be classified into two categories: embedding-based methods [30,31] and generative model-based methods [32–34]. In defect detection cases in industrial scenarios, Zhang et al. [35] proposed the Zero-DD model, which is sensitive to the differential features between normal images and known defect images. The extracted differential features are input into a GAN to discriminate against unknown defect images that fail to meet the requirements, thus achieving zero-shot recognition of unknown defects. So far, literature on ZSL has mainly focused on natural images, ignoring industrial surface defect image detection.

In summary, the majority of deep learning-based defect detection methods currently demonstrate excellent performance under supervised learning. However, some studies have attempted to tackle the few-data problem, with the model's effectiveness being heavily dependent on large amounts of data or relevant field data as auxiliary information, utilizing various data mining techniques for feature recognition. In actual industrial manufacturing processes, the incidence of product defects is low, and defect data is scarce. Furthermore, flexible production lines (order-based production) often involve new products and new processes in small batches. Previous studies [36] have indicated that models pre-trained on other datasets are unable to learn specific feature patterns in certain instances, leading to a reduction in recognition accuracy on other application datasets by 50%~60%. These factors pose a significant challenge for models that rely on big data for feature mining to satisfy the requirements of rapidly changing environments.

In this case, domain experts have prior knowledge of feature data, which is an advantage that machine intelligence cannot match [37]. There have been many studies that have enhanced intelligence through human-machine collaboration, combining the strengths of both to form a new "1 + 1> 2" type of intelligent enhancement, namely, human-machine hybrid intelligence, where information from humans and machines interact and promote each other, and through interaction and collaboration, the performance of artificial intelligence system is enhanced [38], making artificial intelligence a natural extension and expansion of human intelligence, more efficiently solving complex problems [39]. Inspired by the humans-machines hybrid intelligence, this study attempts to involve domain experts in the iterative process of developing deep models, transmitting human experience information to machines, and helping machines to self-reform and iterate through human-machine collaboration, assisting the model in more effectively completing complex and dynamic intelligent tasks.

Based on the analysis above, this study proposes a novel method that complements existing methods, which enhances model intelligence by leveraging human expertise, achieving human-machine hybrid data augmentation in industrial scenarios. Specifically, by intervening based on the expert understanding of exceptional knowledge, data with rich features, such as positions, sizes, and backgrounds can be created, so human knowledge is directly transmitted to the model as prior knowledge bypass the need for big data. The main contributions of this study are as follows:

- A few-data learning method has been proposed, which combines human-machine knowledge to assist in data augmentation. By transferring human knowledge as prior knowledge to the model, making it feasible to accomplish intricate and dynamic tasks in industrial scenarios using few data.
- The method bypasses the need for big data, incorporates domain expert knowledge to guide data creation, and uses style transfer to generate large amounts of high-fidelity data with richness, reducing the model's dependence on big data.
- Verifying the feasibility and effectiveness of the method on an industrial dataset. The experimental results show that the method exhibits outstanding performance in few-data defect detection, outperforming traditional and GAN-based data augmentation methods, providing a novel research method and theoretical support for the existing quality control limitations of manufacturing enterprises.

The rest of this paper is organized as follows. Section 2 introduces related work of industrial defect detection and the mainstream method for limited data. Section 3 introduces the structure of the proposed method. Section 4 conducts various experiments on the magnetic tile dataset. Section 5 shows the conclusion and future work

## 2. Related work

*2.1. Defect detection*

Defect detection refers to the process of identifying and locating relevant defects in images, which is an important task in industrial manufacturing. Before the advent of deep learning methods, traditional approaches [40–43] mainly used handcrafted feature extractors for defect recognition. Some traditional machine learning algorithms [44] were also commonly used. However, these methods require specialized domain knowledge and have limited ability to extract representative features, making them unsuitable for complex industrial scenarios.

With the development of deep learning, more researchers have turned to using convolutional neural network models for defect detection, resulting in significant breakthroughs [45]. Various object detection methods [46,47] and training strategies [48–50] have been developed. For instance, in the study [51], a cross-scale weighted feature fusion network was proposed to identify and locate surface defects in hot-rolled steel, achieving a mean average precision (mAP) of 86.8%. In another study [49], a detection model for intelligent industrial monitoring was proposed to classify and locate multi-scale defects on steel surfaces, achieving an mAP of 80.5%.

Deep learning models rely on large amounts of data, which is difficult to obtain in actual industrial production processes. The yield of defective products is low, resulting in extremely scarce defect data, making the limited defect samples a bottleneck problem [15]. To address this issue, numerous studies have proposed various methods to solve the few-data problem.

*2.2. Learning from few-data*

Defect detection tasks based on deep learning necessitate a considerable number of annotated training samples, making it imperative to obtain a high-quality detector despite limited data. To overcome the challenge of inadequate data, recent research has explored several approaches that can be broadly classified into three categories: data augmentation based on GAN, the FSL method, and the ZSL method.

*2.2.1. Data augmentation based on GAN*

GANs [13] are powerful generative models that train both a generator to generate realistic fake images and a discriminator to distinguish between real and fake images. It excels at generating images with diverse features, making it widely used in data augmentation. By leveraging prior knowledge, GANs learn the true distribution of data and provide rich defect data for subsequent detection. For example, Zhang et al.[52] proposed an automatic defect synthesis network that generates realistic defects in various image backgrounds with different textures and appearances, and flexibly controls the location and category of the defects generated in the image backgrounds. An excellent Defect-GAN was trained on the CODEBRIM1 dataset and an additional 50,000 images. Jain et al.[53] used the generator to synthesize new surface defect images from random noise, which gradually improved over time to become realistic fake images for further training of the classification algorithm. NEU-CLS (1800 images) was used as the dataset for verification. Yang et al. [54] proposed the Mask2Defect network, which injects prior knowledge to generate defects with a large number of different attribute features in a controllable way. The generated defects are used as teacher samples to fine-tune the identification model and achieve classification and localization of defect data. The network was trained using a dataset that includes 1500 images. Other studies have also carried out relevant work [55–57].

However, using GANs to synthesize defect samples still faces some challenges. First, existing GANs still require large-scale training data (at least thousands of images), which has high training costs. Datasets with thousands of images may even consider supervised learning. Second, GANs tend to generate images with feature distributions similar to reference samples, so it is necessary to pass on anomalous data with rich features (such as feature patterns, directions, sizes, positions, backgrounds, etc.). Finally, defects generated by GANs have uncontrollable and limited feature diversity, which cannot fully utilize limited data to fundamentally improve the generalization ability of the defect detection model.

*2.2.2. Few-shot learning method*

FSL refers to completing object recognition tasks with a limited number of labeled samples, and its core is to use prior knowledge to compensate for the lack of supervision information. It can be classified into three categories:

The first category involves using enhanced training data as prior knowledge [20,58,59]. For instance, Tsai et al. [60] proposed a two-stage CycleGAN scheme to automatically synthesize and annotate local defect pixels for the automatic detection of material surface defects. This method eliminates the need for image annotation and model training and achieves the recognition of defect pixels. This method combined with transfer learning to achieve crack detection in a small dataset. However, the drawback of such methods is that the enhancement strategies are usually tailored for each dataset in a particular way for different applications, and the prior data is not easily applicable to other datasets, especially those from different fields.

The second category involves improving the model by limiting the hypothesis space [21,61,62]. For example, Yu et al. [63] proposed a SPNet with a matrix decomposition attention mechanism to solve the segmentation problem of metal surfaces. This method uses matrix decomposition to enhance the salient features of defects and uses a selective prototype module to extract the relationship between images. In [64], a method based on graph embedding and optimal transport was proposed for detecting defects on metal surfaces. The GEDT module uses the correlation information between different features in the dataset to ensure the consistent distribution of graph embeddings, then the OPT module is used to achieve few-shot classification.

The third category involves improving algorithms to find the strategy for searching the best hypothesis within the given hypothesis space [22,65,66]. For example, Song et al. [67] proposed a defect recognition method based on dynamic weighting and joint metrics, which includes an Affine Dynamic Weighting (ADW) module that extracts discriminative features better by learning affine parameters. It also includes a joint metric method that combines the Kullback-Leibler (K-L) divergence covariance measurement module (KLCM) and cosine classifier to optimize and improve the problems of data scarcity and imbalanced defect samples.

However, FSL typically requires a large number of samples from the same or related domains, and still requires the establishment of a large amount of universal semantic datasets, while also detailed annotation of abnormal sample attributes. The existing dataset bias is also one of the urgent problems to be solved at present. Therefore, its performance is quite limited and still far from practical application in the industrial field.

*2.2.3. Zero-shot learning*

ZSL is a method that utilizes known category information or sides common knowledge information as attribute labels to train models [68,69]. Through sharing semantics, ZSL can achieve recognition of unknown categories, solve the problem of lacking category labels, and even perform classification when the training set and testing set are not intersected [28,29]. ZSL can be classified into two categories: embedding-based methods and generative model-based methods.

The first category is based on embedding methods [30,31,70]. A projection function is used to map image features and semantic attributes to a common embedding space [71], and deep learning models are trained with known class information or side common knowledge information as attribute labels to achieve recognition of unknown classes through sharing semantics. For example, Li et al. [72] proposed the Class Knowledge Graph Construction (CKGC) method to establish relationships between basic defect classes and new defect classes, then used graph convolutional neural networks to learn class features, and finally inputted image features and class features into the classifier to predict defect class information.

The second category is based on generative models [40-42], aiming to generate image features for unseen classes using semantic attributes, usually accomplished by conditional generative adversarial networks (CGANs). Currently, this method lacks practical examples in the field of industrial defect detection, but there are already studies that have used this method to complete relevant applications. For example, Xian et al. [73] constructed a generative model using a GAN, where the discriminator is trained with the classification loss of class attributes, and the generator is based on the conditional GAN to generate visual features corresponding to class attributes.

The limitation of ZSL is that the model still requires a large amount of auxiliary information for training, and the required auxiliary information attributes are usually defined manually, resulting in high annotation costs. In addition, there are bias and domain shift problems, and the model tends to predict seen class labels as output, which makes it difficult to adapt and generalize well to unseen classes outside the distribution.

In summary, the existing methods for addressing few-data problems all share a common characteristic, which is their reliance on a large amount of data or related domain data as auxiliary information, and then using various data mining methods to achieve defect recognition. The intelligence of these models relies explicitly on model training tens of thousands of times on large datasets. However, these models are prone to catastrophic failures on new datasets. The challenge is to enable models to adapt to changing tasks, and our method is a feasible solution that starts from the data level and utilizes human intelligence to assist models to become more intelligent. By using a human-machine hybrid intelligent data augmentation method, we reduce the model's dependence on existing data, making it possible to achieve defect detection using a few data in industrial scenarios.

## 3. Methodology

In this section, we introduce our proposed human-machine knowledge hybrid data augmentation method (Fig. 2), which integrates expert knowledge to guide the model in creating images with rich features from the data level, realizing anomaly detection in few-data scenarios. The method consists of four stages: image editing, style transfer, image filtering, and image classification.

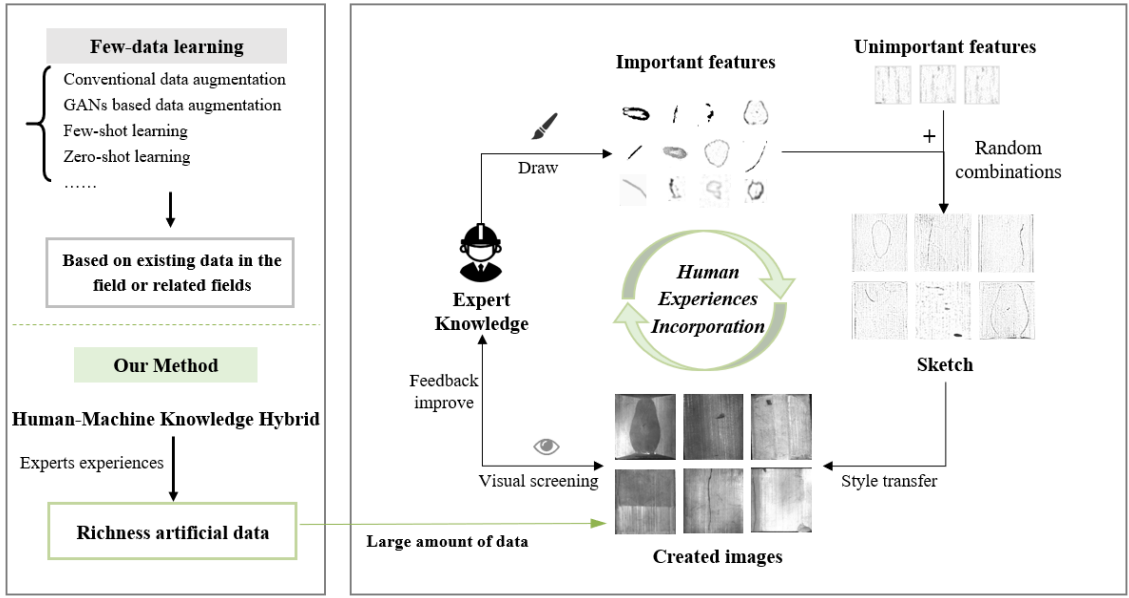

Fig. 2. Overall framework of the proposed method

**Step. 1 Image Editing.**

To address the challenge of extremely sparse or non-existent few-data situations in industrial defect detection, this study utilizes the expert's understanding of anomaly knowledge as an imaging method to create richness dataset from scratch. However, this method faces two challenging conditional limitations. Firstly, expecting experts to provide hundreds or thousands of defect features is unreasonable, so the expert creation process of anomaly samples needs to be simplified. Secondly, the images provided by the expert only contain sketches of the main defects in the target domain and lack realism. Therefore, synthesizing high-fidelity images by combining real attribute features with the expert's sketches is essential, to create a more realistic industrial dataset.

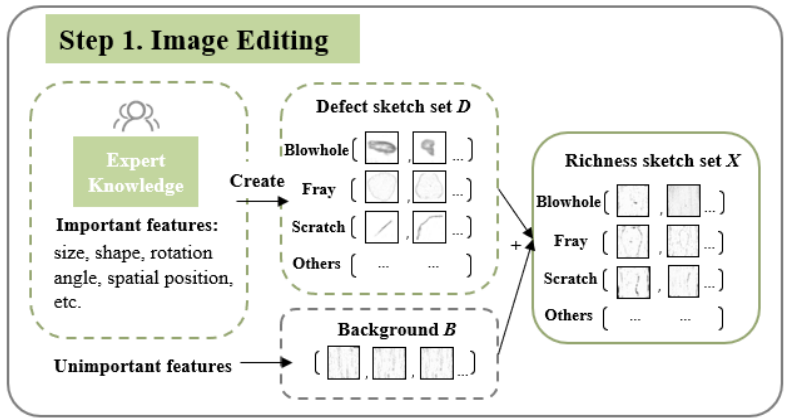

Fig.3 Flowchart of image editing (Step.1).

To address these issues, we propose an image editing method (Fig. 3) as follows: First, domain experts identify important defect features such as size, shape, rotation angle, and spatial position to create a collection of possible defect sketches, denoted as $D$. Next, we combine important features from $D$ and unimportant background features from $B$ randomly through manual methods to create a large collection of synthetic sketches $X$ with rich features, positions, sizes, and backgrounds (Fig. 4). This method not only incorporates expert knowledge guidance but also generates a large number of diverse images with different attribute features using a small number of synthetic sketches, which improves the recall of defect-recognition in the experimental results.

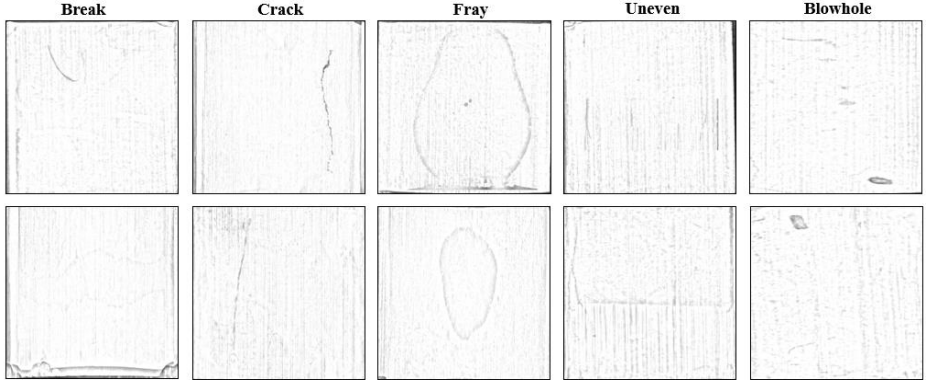

Fig.4 Example diagram of artificial sketch set $X$ in different categories

| Algorithm 1: Image editing. |
|---|
| **Input:** defective region set $D_o$ drawn by the expert, background image set $B$, x-axis scaling coefficient $p$, y-axis scaling coefficient $q$, rotation angle $\theta$, the center coordinates $(x_o, y_o)$ |
| **Output** sketch set $X$ |
| 1: **do** GaussianBlur and Grayscale to get sketch set $D$ |
| 2: **do** data augmentation in $D$ and $B$: |
| 3:   clip, crop, zoom, translation, rotation |
| 4: **return** $D$, $B$ |
| 5: **for** $D_i$ in $D$ |
| 6:   **do** Binarization to get a mask $M_i$ |
| 7: **return** $M$ |
| 8: initialize $h, j, (x_o, y_o), \theta, p, q$ |
| 9: choose $D_i$ as the defective features, $B_i$ as the background |
| 10: **do** $D_i$, $M_i$ rotate with the angle of $\theta$, x direction, and y direction are scaled by coefficients $p, q$ |
| 11: **fuse** $D_i$ on $B_i$ with center point $(x_o, y_o)$ to get $X_i$ |
| 12:   **for** each pixel value in the mask $M_i$: |
| 13:     **if** the pixel value = 1: |
| 14:       the pixel value of $X_i = D_i$ |
| 15:     **else:** |
| 16:       the pixel value of $X_i = B_i$ |
| 17:   **end** |
| 18: **end** |
| 19: **return** sketch image set $X$ |

**Step. 2 Style transfer.**

To generate realistic images from expert sketches, this study employs the domain adaptation method, with expert knowledge as the generated content and real images as the generation style, to construct a domain transfer adversarial model (Fig. 5). The model is guided by sketches to generate high-fidelity images with expert knowledge. The goal of the domain transfer adversarial model is to learn the mapping function between two domains $X$ and $Y$, where $\{x_i\}_{i=1}^{N}, x_i \in X$, and $x \sim P_{data}(x)$ represents the sketches containing main features, while $\{y_i\}_{j=1}^{M}, y_j \in Y$, and $y \sim P_{data}(y)$ represents the real images containing some irrelevant features and the image generation style that we need. The model includes two mappings $G: X \rightarrow Y$ and $F: Y \rightarrow X$. Mapping G ensures the distribution matching of the original and target domains, and mapping F constrains the high consistency between the original and target domain data. After obtaining a well-trained model, the artificially created sketches set $X$ is input to generate the real images $Y'$. Through this style transfer method, we successfully transformed expert sketches into high-fidelity images with the style of the true images, as shown in Fig. 6. With more data used to train the network, more realistic image effects can be obtained.

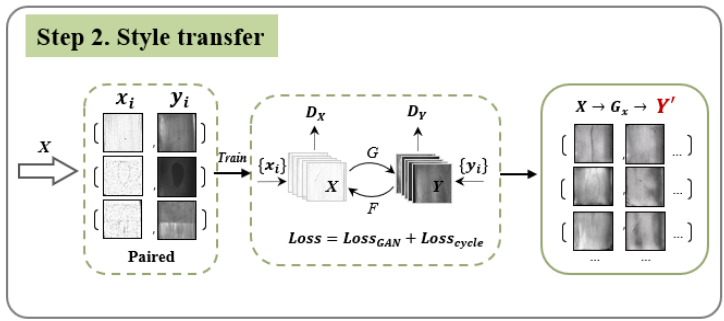

Fig.5 Flowchart of style transfer (Step.2)

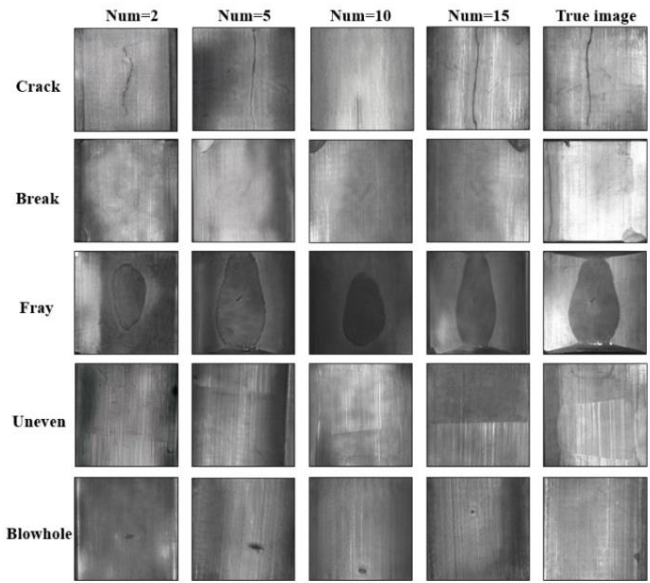

Fig.6 Example diagram of the generated image set $Y'$, using 2, 5, 10, and 15 training images respectively.

### Step. 3 Image filtering.

To ensure that the large randomly generated dataset matches the real images and does not negatively impact the subsequent classification model, an image screening step was introduced (Fig. 7). The image screening process involves visual inspection and t-SNE analysis. The visual inspection method is used to roughly filter out examples where the defect features do not match the actual geometry in space, and to filter out visually dissimilar images. For images that cannot be filtered out manually, the t-SNE method is used for screening. This method maps high-dimensional data to a low-dimensional space and represents the similarity between data points using t-distribution. By transforming the similarity between data points into conditional probability, it helps to understand the data structure while retaining the local characteristics of the dataset. The defect data of each category is visualized, and a distance threshold value $n$ is set between the generated image $Y'$, and the real image $X$. Images with a threshold greater than $n$ are filtered out to obtain the final usable dataset. The subsequent experimental results demonstrate that the model's recognition performance has indeed improved after the screening process.

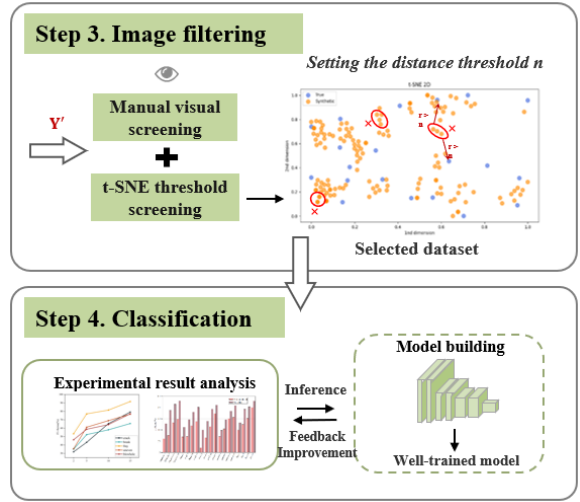

Fig.7 Flowchart of image filtering and classification (Step.3 and Step.4)

---

**Algorithm 2:** Image filtering.

**Input:** created image set $X = \{x_1, x_2, \ldots, x_n\}$, true image set $X' = \{x'_1, x'_2, \ldots x'_m\}$, threshold $N$, cost function parameters: perplexity $Perp$, optimization parameters: number of iterations $T$, learning rate $\eta$, momentum $\alpha(t)$.

**Output:** image set $Y$ after image filtering

1: **for** each image in $X$ and $X'$:

2: compute pairwise affinities $p_{j|i}$ with perplexity $Prep$: $p_{j|i} = \frac{exp(-||x_i-x_j||^2/2\delta_i^2)}{\sum_{k \neq i} exp(-||x_i-x_k||^2/2\delta_i^2)}$

3: set $p_{ij} = \frac{p_{j|i}+p_{i|j}}{2n}$

4: sample initial solution $Y^{(t)} = \{y_1, y_2, \ldots, y_n\}$ from $N(0, 10^{-4}i)$

5: **for** $t=1$ to T **do**

6:    compute low-dimensional affinities $q_{ij}$: $q_{ij} = \frac{(1+||y_i-y_j||^2)^{-1}}{\sum_{k \neq l}(1+||y_k-y_l||^2)^{-1}}$

7:    compute gradient $\frac{\delta C}{\delta y_1}$ using: $\frac{\delta C}{\delta y_1} = 4 \sum_j (p_{ij} - q_{ij})(y_i - y_j)(1 + ||y_i - y_j||^2)^{-1}$

| | |
|---|---|
| 8: | set $Y^{(t)} = Y^{(t-1)} + \eta \frac{\delta C}{\delta y} + \alpha(t)(Y^{(t-1)} - Y^{(t-2)})$ |
| 9: | end |
| 10: | **return** $Y = \{y_1, y_2, \ldots, y_n\}$, $Y' = \{y'_1, y'_2, \ldots y'_m\}$ |
| 11: | for i in n: |
| 12: |   for j in m: |
| 13: |     computes the Euclidean distance $n$ between $y_i$ and $y'_j$ |
| 14: |     if n > N: |
| 15: |       filter out $y_i$ from $Y$ |
| 16: |     end |
| 17: |   end |
| 18: | return $Y$ |

**Step. 4 Image classification.**

Finally, the dataset with expert knowledge is input into a defect detection model for training. For the selection of the classification network, we conducted experiments using VGG, ResNet, and MobileNet networks. The classification results among the networks showed no significant difference. However, the VGG network demonstrated superior classification performance for the task in this study. Therefore, we ultimately chose VGG16 as an example to showcase the results. The VGG16 consists of five convolutional layers, three fully connected layers, and a softmax output layer. Max-pooling is used to separate the layers, and the activation units of all hidden layers use ReLU functions. We use pre-trained weights for training, and after obtaining a well-trained model, we test it on the same test set and make appropriate feedback adjustments based on the experimental results.

## 4. Experimental Results and Analysis

This section validates the proposed method through case studies. Firstly, we introduced the dataset and evaluation metrics used in the study. Then, the proposed method is used to create data, followed by verification experiments, comparative experiments of different methods, and multi-classification experiments. Finally, we discussed the experiment results and analyze the limitations of the proposed method.

*4.1. Dataset and metrics*

Experimental analysis of the proposed method was performed using the Magnetic Tile Dataset (MTD), which comprises cropped ROI images of magnetic tile surfaces. The dataset aims to detect various surface defects, including six categories: Blowhole, Crack, Fray, Break, Uneven (caused by the grinding process), and Free (non-defective) (Fig. 8). The dataset contains a total of 1344 images, with 392 defective images and 952 non-defective images, each with a size of 300 x 300 pixels. The division of the dataset is shown in Table 1. The experiments were conducted using the PyTorch framework of Python3.7, on a server equipped with a size of 300 x 300 pixels. The experiments were conducted using the PyTorch framework of Python3.7, on a server equipped with an Intel Xeon Silver 4210R CPU and NVIDIA Geforce RTX 3090 GPU.

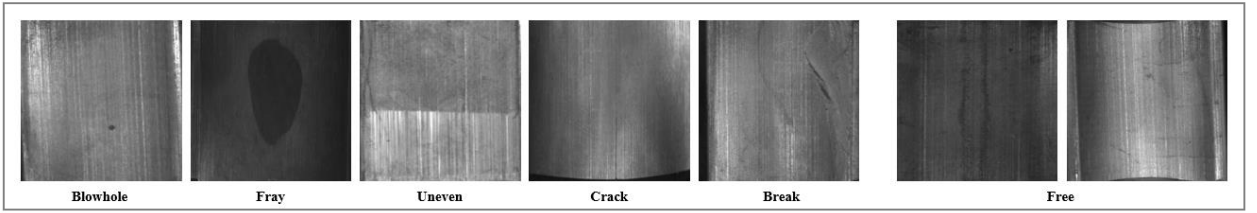

Fig.8 Images of each category in the Magnetic Tile Dataset.

Table 1. Division of Magnetic Tile Dataset in experiments.

| Category | Blowhole | Fray | Uneven | Crack | Break | Free |
|---|---|---|---|---|---|---|
| Train | 2-15 | 2-15 | 2-15 | 2-15 | 2-15 | 152 |
| Test | 50 | 12 | 30 | 20 | 30 | 800 |
| Total | 115 | 32 | 103 | 57 | 85 | 952 |

In the experiments, the evaluation indicators used to assess the classification performance are Accuracy, Recall, Precision, and F1-score. Accuracy measures the proportion of correctly predicted samples, encompassing all categories: $Accuracy = (TP + TN)/(TP + FP + TN + FN)$. Recall reflects the model's ability to correctly predict true defect images: $Recall = TP/(TP + FN)$. Precision measures the proportion of defect predictions that are truly defects: $Precision = TP/(TP + FP)$. F1-score is a balanced mean of Recall and Precision that can be used to assess the overall performance of the model: $F1\ score = 2 \times Recall \times Precision/(Recall + Precision)$.

*4.2. Experimental results*

*4.2.1. Image generation*

Due to the scarcity of defective data and the rarity of some defects, the experiment attempted to transfer expert knowledge of abnormality to the model as prior knowledge. Firstly, the domain expert drew dozens of defect area sketches $D$ for five types of defects and obtained $D'$ after a simple expansion of $D$. Then, $D'$ was fused with both the defective and non-defective background. Based on expert experience, some images that conform to the characteristics of magnetic tiles were screened out, and the artificial sketch set $X$ was obtained (Fig. 4). The specific number of images for each step is shown in Table 2.

Table 2. The specific number of images in the image editing process for each step. $D$: Sketch set of the defective area drawn by an expert. $D'$: Sketch expansion. $X$: The sketches set after merging defects into the background

| Category | Crack | Break | Fray | Uneven | Blowhole |
|---|---|---|---|---|---|
| $D$ | 37 | 55 | 20 | 70 | 60 |
| $D'$ | 74 | 110 | 80 | 100 | 100 |
| $X$ | 300 | 260 | 230 | 350 | 300 |

To generate images with the style of the magnetic tile using sketch set $X$, a domain transformation adversarial model was created and trained in this section. This model guides the shape and location of defects in real images by sketches and generates high-fidelity images with expert knowledge. The domain-transformed adversarial model was trained with 2, 5, 10, and 15 images from each category, after data augmentation. Then, the sketch set $X$ was fed into the well-trained domain-transformation adversarial model to obtain the generated image set $Y'$. As shown in Fig. 6, the generated images have a high similarity with real magnetic tile images, and the defect features drawn by the expert are visible in the visualization. However, the detailed defect features (such as color and shape) are different from the real defects. With the addition of training data, the model's performance improved, resulting in more realistic defect features with detailed edge features, and overall uniform images without large abnormal features due to local effects.

Furthermore, the method of manually visualizing and setting a threshold with t-SNE was used to filter out images with low similarity. In this process, a closed-loop feedback method was also employed, and based on the results of the testing set, negative feedback data was repeatedly filtered to ensure that the screening could gradually improve the model's performance. The experimental results showed that this step did indeed increase the final classification accuracy. The number of images in the image generation and image screening stages can be found in Table 3.

Table 3. The number of images in each category at different stages. $Y'$: generate images set. $M$: Screened dataset for classification.

| Category | Crack | | | | Break | | | | Fray | | | |
|---|---|---|---|---|---|---|---|---|---|---|---|---|
| Training set | 2 | 5 | 10 | 15 | 2 | 5 | 10 | 15 | 2 | 5 | 10 | 15 |
| $Y'$ | 300 | 300 | 300 | 300 | 260 | 260 | 260 | 260 | 230 | 230 | 230 | 230 |
| $M$ | 224 | 193 | 324 | 374 | 215 | 220 | 250 | 265 | 180 | 185 | 205 | 300 |
| Category | Uneven | | | | Blowhole | | | | Free | | | |
| Training set | 2 | 5 | 10 | 15 | 2 | 5 | 10 | 15 | - | | | |
| $Y'$ | 350 | 350 | 350 | 350 | 300 | 300 | 300 | 300 | - | | | |
| $M$ | 186 | 203 | 346 | 349 | 230 | 230 | 200 | 279 | 152 | | | |

### 4.2.2. Classification

In this section, the goal is to achieve final defect detection using the dataset obtained in the previous section for binary classification. The main evaluation metrics include Accuracy, Recall, Precision, and F1-score. Since it is more critical to correctly detect defective products than qualified products in industrial production, the experiments' recall and precision rates are only for the defect category. A pre-trained Vgg16 is utilized as the classification network with the Adam optimizer, and training epochs range from 50 to 200.

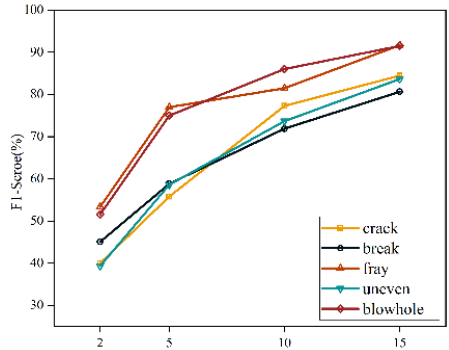

Fig. 9 The F1-score of different categories when the number of training images increases.

Table 4. Defect detection results for different category images.

| Category | Crack | | | | Break | | | | Fray | | | |
|---|---|---|---|---|---|---|---|---|---|---|---|---|
| Training set | 2 | 5 | 10 | 15 | 2 | 5 | 10 | 15 | 2 | 5 | 10 | 15 |
| Recall | 45.00 | 60.00 | 85.00 | 95.00 | 53.33 | 66.67 | 76.67 | 83.33 | 66.67 | 82.71 | 91.67 | 91.67 |
| Precision | 36.00 | 52.17 | 70.83 | 76.00 | 39.02 | 52.63 | 67.65 | 78.13 | 44.44 | 71.97 | 73.33 | 91.67 |
| F1 | 40.00 | 55.81 | 77.27 | **84.44** | 45.07 | 58.82 | 71.87 | **80.65** | 53.33 | 76.98 | 81.48 | **91.69** |
| Category | | Uneven | | | | Blowhole | | | | | | |
| Training set | 2 | 5 | 10 | 15 | 2 | 5 | 10 | 15 | | | | |
| Recall | 40.00 | 56.67 | 70.00 | 76.67 | 50.00 | 66.00 | 80.00 | 86.00 | | | | |
| Precision | 38.17 | 60.71 | 77.78 | 92.00 | 53.19 | 86.84 | 93.02 | 97.73 | | | | |
| F1 | 39.34 | 58.62 | 73.69 | **83.64** | 51.55 | 75.00 | 86.02 | **91.49** | | | | |

Fig. 9 illustrates the F1-score of the method with an increasing number of training images. The detailed experimental results for each category are shown in Table 4. The experimental results indicate that the proposed method can train a model with good classification performance even when defect data is extremely rare or non-existent. With the increase of training images, the model's classification performance on defect images also improves. Conversely, the "fray" category exhibits the highest improvement, with the F1-score increasing to 91.69% when 15 images are used for training. This result may be due to the relatively fewer defect types in this category and the prior knowledge provided by experts, which can encompass most of the possible defect types, resulting in a better performance on the testing set. The "break" category has relatively fewer improvements, with the F1-score only reaching 80.65% when 15 images are used. It is speculated there are many types of defects in the "break" category, and the characteristics have significant differences, making it difficult to describe the exact characteristics using 2D sketches. Therefore, it requires more labor costs for correction and selection than other categories. Additionally, the performance of the proposed method varies across different defect categories, with different degrees of improvement observed. The performance and improvement degree is related to the characteristics of the defects themselves and the amount of expert knowledge provided. Therefore, a detailed expert knowledge intervention is required to address the specific defect scenarios in real-world applications and obtain better detection results. The confusion matrix of the experiments is shown in Fig. 10.

Fig.10. Classification confusion matrix for each defect category under different numbers of training images

*4.2.3. Comparison with different methods*

To further evaluate the performance of the proposed method, this section compares the classification performance with other mainstream methods, including traditional augmentation methods and generative models.

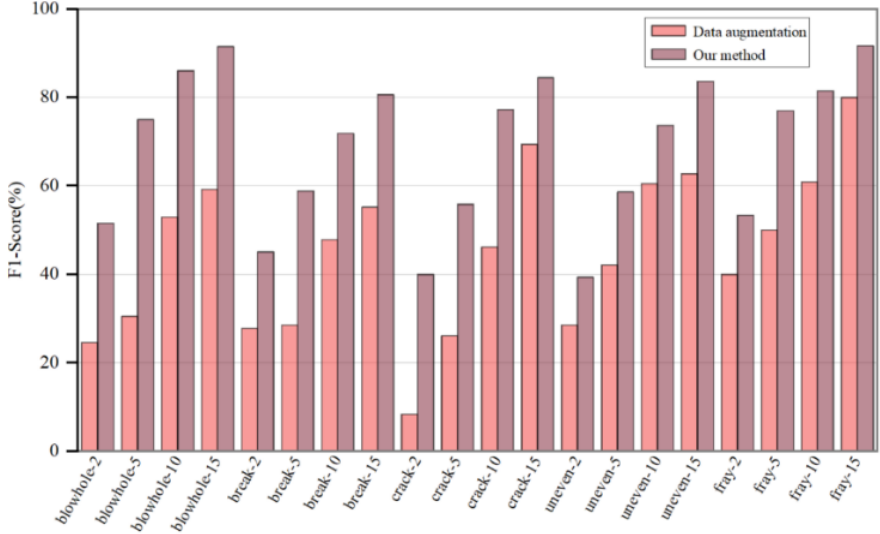

Fig. 11. F1-score comparison between traditional augmentation methods and our method under different types of defects

We conducted comparative experiments by applying different methods using 2, 5, 10, and 15 images for each category and then feeding them into the same classification network for defect recognition. The experimental results are presented in Table 5. The recall and precision values reported in the results are only for the defect categories, and the F1-score of the comparative experiments are shown in Fig.11. The results indicate that our proposed method improves the recognition ability of the model based on traditional augmentation methods, with the degree of improvement varying depending on the category. Notably, our method achieves higher improvement for categories with distinct defect features (e.g., crack and fray), while the effect is slightly worse for categories with complex and fuzzy defect features. Furthermore, we observed that traditional augmentation methods often lead to models misclassifying images as defect-free categories (i.e., the precision value is abnormally high), indicating poor model training. This issue is particularly disadvantageous in industrial scenarios where the model must be sensitive to abnormal data. However, increasing the training data can alleviate this problem. In our proposed method, abnormal images with rich features effectively resolve this problem.

Table 5. Comparison results between traditional augmentation methods and our method under different types of defects

| **Blowhole** | Test set (defect/free) | Data augmentation | | | Our method | | |
|---|---|---|---|---|---|---|---|
| | | Recall | Precision | F1 | Recall | Precision | F1 |
| 2 | 50/800 | 14.00 | 100.00 | 24.56 | 50.00 | 53.19 | 51.55 |
| 5 | 50/800 | 18.00 | 100.00 | 30.51 | 66.00 | 86.84 | 75.00 |
| 10 | 50/800 | 36.00 | 100.00 | 52.94 | 80.00 | 93.02 | 86.02 |
| 15 | 50/800 | 58.00 | 60.42 | 59.18 | 86.00 | 97.73 | **91.49** |
| **Break** | Test set (defect/free) | Data augmentation | | | Our method | | |
| | | Recall | Precision | F1 | Recall | Precision | F1 |

| | 2 | 30/800 | 16.67 | 83.33 | 27.78 | 53.33 | 39.02 | 45.07 |
| | 5 | 30/800 | 20.00 | 50.00 | 28.57 | 66.67 | 52.63 | 58.82 |
| | 10 | 30/800 | 36.67 | 68.75 | 47.83 | 76.67 | 67.65 | 71.88 |
| | 15 | 30/800 | 53.33 | 57.14 | 55.17 | 83.33 | 78.13 | **80.65** |

| **Crack** | Test set (defect/free) | Data augmentation | | | Our method | | |
|---|---|---|---|---|---|---|---|
| | | Recall | Precision | F1 | Recall | Precision | F1 |
| 2 | 20/800 | 5.00 | 25.00 | 8.33 | 45.00 | 36.00 | 40.00 |
| 5 | 20/800 | 15.00 | 100.00 | 26.09 | 60.00 | 52.17 | 55.81 |
| 10 | 20/800 | 30.00 | 100.00 | 46.15 | 85.00 | 70.83 | 77.27 |
| 15 | 20/800 | 85.00 | 58.62 | 69.39 | 95.00 | 76.00 | **84.44** |

| **Uneven** | Test set (defect/free) | Data augmentation | | | Our method | | |
|---|---|---|---|---|---|---|---|
| | | Recall | Precision | F1 | Recall | Precision | F1 |
| 2 | 30/800 | 16.67 | 100.00 | 28.57 | 40.00 | 38.71 | 39.34 |
| 5 | 30/800 | 26.67 | 100.00 | 42.11 | 56.67 | 60.71 | 58.62 |
| 10 | 30/800 | 43.33 | 100.00 | 60.47 | 70.00 | 77.78 | 73.69 |
| 15 | 30/800 | 53.33 | 76.19 | 62.75 | 76.67 | 92.00 | **83.64** |

| **Fray** | Test set (defect/free) | Data augmentation | | | Our method | | |
|---|---|---|---|---|---|---|---|
| | | Recall | Precision | F1 | Recall | Precision | F1 |
| 2 | 12/800 | 25.00 | 100.00 | 40.00 | 66.67 | 44.44 | 53.33 |
| 5 | 12/800 | 33.33 | 100.00 | 50.00 | 82.71 | 71.97 | 76.98 |
| 10 | 12/800 | 58.33 | 63.64 | 60.87 | 91.67 | 73.33 | 81.48 |
| 15 | 12/800 | 66.67 | 100.00 | 80.00 | 91.67 | 91.67 | **91.69** |

In our comparative experiments with generative models, we evaluated the performance of DCGAN [74], the traditional data augmentation method, and our proposed method. Specifically, we focused on the "uneven" category and varied the number of training images (2, 5, 10, and 15) to assess the effectiveness of each method. We present the images synthesized by DCGAN in Fig. 12, which demonstrate that using a generative model to create defect images is ineffective when facing extremely limited data conditions. This is because GANs are unable to generate images with diverse features in such scenarios. Although using the entire dataset for training improves the richness of the generated images, their realism is poor. In contrast, our proposed method generates realistic and diverse images using only a few images, making it more appropriate for scenarios with few data.

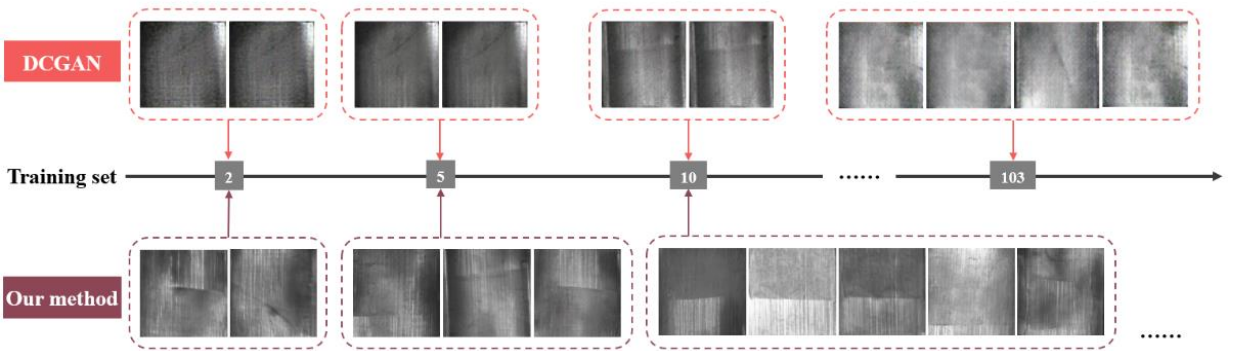

Fig. 12. Images generated by DCGAN and our method when using different numbers of images as the training set

We then fed the enhanced data into the classification network and present the results in Table 6. The F1 values for detection are summarized in Fig. 13, which shows that the use of GAN for data augmentation did not improve the detection performance of the model. The unrealistic defect images generated by GAN even reduced the performance of the classification model, leading to lower overall results than the traditional image augmentation method. Furthermore, models trained using GAN-based data augmentation tended to predict non-defective categories, further demonstrating that GAN cannot improve the classification performance of the model by generating samples with diverse features under few data conditions.

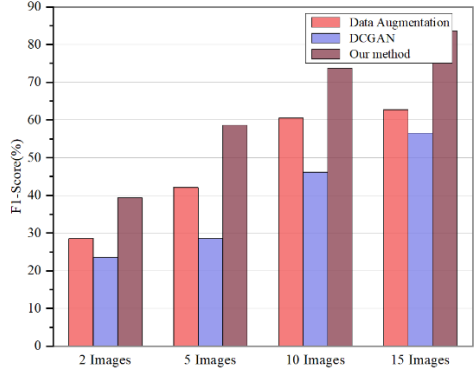

Fig. 13. F1-score comparison using DCGAN, data augmentation, and our method

Table 6. Comparative experimental results using DCGAN, data augmentation, and our method.

| Metrics | Data augmentation | | | DCGAN | | | Our method | | |
| --- | --- | --- | --- | --- | --- | --- | --- | --- | --- |
| (%) | Recall | Precision | F1 | Recall | Precision | F1 | Recall | Precision | F1 |
| 2 | 16.67 | 100.00 | 28.57 | 13.33 | 100.00 | 23.53 | 40.00 | 38.71 | 39.34 |
| 5 | 26.67 | 100.00 | 42.11 | 16.67 | 100.00 | 28.57 | 56.67 | 60.71 | 58.62 |
| 10 | 43.33 | 100.00 | 60.47 | 30.00 | 100.00 | 46.15 | 70.00 | 77.78 | 73.69 |
| 15 | 53.33 | 76.19 | 62.75 | 43.33 | 81.25 | 56.52 | 76.67 | 92.00 | **83.64** |

*4.2.4. Multi-classification experiments*

In practical production, depending on the specific quality control requirements of the product, sometimes classifying multiple defect types is necessary. Therefore, this section evaluates the performance of the proposed method on multi-classification tasks. Multi-classification tasks can be divided into two types: one is to treat all defect types as a unified nonconforming category for detection, and the other is to identify the type of each defect separately. We conducted comparative experiments on these two multi-classification scenarios, using original data and traditional augmentation data as controls.

We first conducted experiments by treating all defect types as unqualified categories. Table 7 displays the classification results obtained using different enhancement methods, and the recall and F1 scores are summarized in Fig. 14. Our proposed method achieved an F1-score of 60.73% using only 2 training images per category, while the F1-scores for the data augmentation method and the original data were 40% and 10.67%, respectively. These experimental results indicate that incorporating defect features as prior knowledge is an effective approach to improving the multi-type defect recognition model. When the number of images increased to 15, the F1-score of the model reached 82.81%, with only a minor overall improvement, which could be due to the limited artificially added defect features. Therefore, when the amount of data increases, the effect of this method becomes less significant. In future work, it is essential to enrich the prior knowledge by manually editing more diverse features to obtain a better detection model.

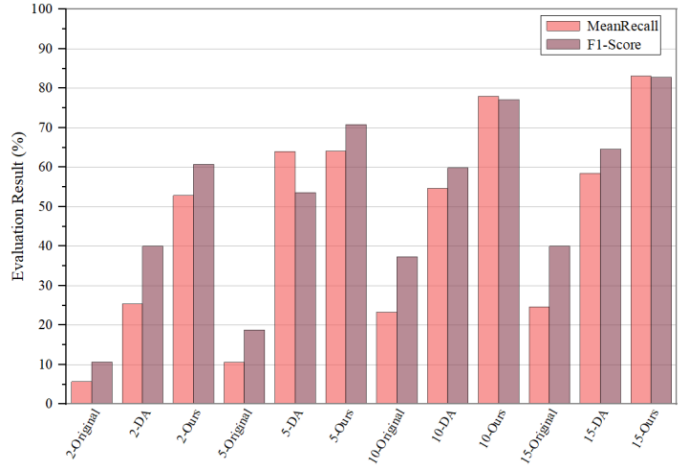

Fig. 14 Comparison of Recall and F1-score using different methods for multi-classification

Table 7. Multi-classification results using different methods

| Metrics (%) | Original dataset | | | Data augmentation | | | Our method | | |
|---|---|---|---|---|---|---|---|---|---|
| | Recall | Precision | F1 | Recall | Precision | F1 | Recall | Precision | F1 |
| 2 | 5.63 | 100.00 | 10.67 | 25.35 | 94.74 | 40.00 | 52.82 | 71.43 | **60.73** |
| 5 | 10.56 | 83.33 | 18.75 | 64.00 | 45.96 | 53.53 | 64.08 | 79.13 | **70.82** |
| 10 | 23.24 | 94.29 | 37.29 | 66.20 | 54.65 | 59.87 | 78.17 | 76.03 | **77.09** |
| 15 | 24.64 | 97.22 | 39.33 | 58.45 | 72.17 | 64.59 | 83.10 | 82.52 | **82.81** |

We proceeded to classify the six defect categories in the dataset, and the results of different methods using varying numbers of training images are shown in Table 8. When only two images were used for each category, our method achieved an F1-score of 52.88%, which increased to 68.07% when the number of images was increased to 15, resulting in an improvement of 11.44% compared to traditional data augmentation methods (Fig.15). Nonetheless, the overall improvement was relatively modest compared to the defect-non-defect detection task, suggesting that there is still significant potential for improvement in the multi-class defect classification method.

Table 8. Multi-classification (6 categories) results using different methods. MR: Mean Recall. MP: Mean Precision

| Metrics (%) | Original dataset | | | Data augmentation | | | Our method | | |
|---|---|---|---|---|---|---|---|---|---|
| | MR | MP | F1 | MR | MP | F1 | MR | MP | F1 |
| 2 | 20.83 | 30.87 | 24.87 | 34.22 | 81.08 | 48.13 | 42.36 | 70.35 | **52.88** |
| 5 | 26.39 | 30.93 | 28.47 | 43.94 | 65.94 | 52.74 | 47.83 | 71.52 | **57.33** |
| 10 | 30.56 | 47.69 | 37.24 | 50.81 | 59.82 | 54.94 | 52.26 | 75.29 | **61.69** |
| 15 | 31.44 | 81.05 | 45.31 | 51.82 | 62.44 | 56.63 | 75.03 | 62.29 | **68.07** |

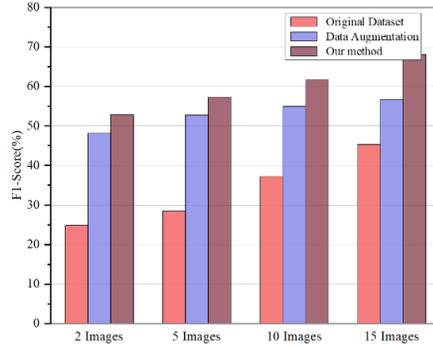
Fig.15 F1-score comparison of multiple classifications using different methods

*4.3. Discussion*

In this section, the experiments consist of three parts: validation of the proposed method, comparison experiments of different methods, and multi-class experiments. Firstly, the MTD dataset is used to verify the performance of the proposed method, and classification experiments are conducted on each category using 2, 5, 10, and 15 images respectively. The experimental results show that the proposed method can achieve good classification performance even when defect data is extremely rare or non-existent. The model's classification performance improves with the increase of training images, but the degree of performance improvement is related to the defect features and the amount of expert knowledge provided, requiring some manual intervention. Among all categories, the "fray" category performs the best, achieving an F1-score of 53.33% with two training images and improving to 91.69% with 15 images.

We conducted a series of experiments to compare the effectiveness of traditional data augmentation methods and generative models. The results revealed that for handling extremely scarce data, complex generative models were unable to generate a diverse range of defect images. Furthermore, the highly distorted generated images often led to poor classification performance. Although traditional data augmentation methods outperformed generative models, they still had a low recall value, frequently classifying defective images as non-defective. This shortcoming could hamper the model's sensitivity and generalization to abnormal data in industrial scenarios where new practical information is scarce. In contrast, our proposed method incorporates prior knowledge into the model and artificially adds various types of defect features, regardless of whether they are similar to real defects. As a result, the model can identify defective images more accurately and achieve better performance.

Finally, we conducted multi-class experiments and compared our method with different approaches. When only detecting the presence or absence of defects, our method achieved F1 scores of 60.73%, 70.82%, 77.09%, and 82.81% with 2, 5, 10, and 15 training images, respectively. Compared to traditional data augmentation methods, our method showed better performance, confirming the feasibility of using manual editing as prior knowledge for the model. In the six-class experiment, our method achieved an F1 score of 68.07% with 15 training images, with a small improvement margin, which is one of the optimization tasks for future work.

To sum up the above experiments, the method proposed in this study still has the following limitations: First, in this work, the added defect features through manual editing are limited, which results in a low improvement in the end, and it is difficult to meet the defect detection standards required for actual industrial implementation. Collecting more defect features and generating more unpredictable defect types through them is the optimization work we need to do in the future. Secondly, there are many manual intervention steps in the whole experiment process (sketch generation, style transfer, etc.). Therefore, simplifying the experiment process, reducing cumulative errors, and achieving an end-to-end detection model are issues that need to be addressed. Finally, this method is not effective in detecting multiple types of defects at the same time, the model is not sensitive enough to different types of features, and the generalization is poor, so the model performance needs to be improved at each stage.

## 5. Conclusions and outlooks

This study proposes a human-machine hybrid intelligence method to address the extremely scarce data problem in industrial applications caused by the rapidly changing task demand (such as multi-variety and small-batch business models). The expert knowledge of abnormal data is directly transferred to the model as prior knowledge, enabling the defect detection task to be achieved in few-data situations. The method can quickly accumulate data from scratch and reduce the model's dependence on big data by utilizing human-machine collaboration to assist the model in rapid iteration and innovation, thereby enabling more effective completion of complex dynamic intelligent tasks in industrial scenarios.

The contributions of this study are as follows: first, we propose a few-data learning method that combines human-machine knowledge to train deep models, enabling the completion of complex dynamic tasks in industrial scenarios with few-data. Second, our method bypasses the need for big data and generates high-fidelity data with rich features, such as position, size, and background, from scratch using expert knowledge of anomalies, reducing the model's dependence on large amounts of data. Third, our experimental results on industrial datasets demonstrate that the proposed method outperforms traditional and GAN-based data augmentation methods in few-data industrial defect detection. This finding provides a new research method and theoretical support for overcoming quality control limitations in manufacturing enterprises.

In future work, we plan to conduct more comprehensive research on our proposed method. Specifically, we will focus on (1) optimizing the multi-category detection results to improve the model's sensitivity and generalization to category features, and enhance its detection performance at each stage. (2) Simplifying the experimental process by reducing the number of tedious human intervention steps, and ultimately realizing an end-to-end detection process. (3) Exploring methods for defect detection tasks that require no data, thus eliminating the model's dependence on data and relying solely on human knowledge to accomplish recognition tasks in zero-data learning. These future research directions will further advance our proposed approach and contribute to the field of industrial defect detection.